\documentclass{article}

\PassOptionsToPackage{numbers, compress}{natbib}

\usepackage{multirow}
\usepackage{graphicx}
\usepackage{booktabs}
\usepackage{colortbl}
\usepackage{amsmath}

\usepackage[table,xcdraw]{xcolor}

\usepackage[preprint]{neurips_2025}

\usepackage[utf8]{inputenc} 
\usepackage[T1]{fontenc}    
\usepackage{hyperref}       
\usepackage{url}            
\usepackage{booktabs}       
\usepackage{amsfonts}       
\usepackage{nicefrac}       
\usepackage{microtype}      
\usepackage{xcolor}         
\usepackage{pifont}
\usepackage{amssymb}
\usepackage{wasysym}
\usepackage{caption}
\usepackage{subcaption}
\usepackage{hyperref} 

\title{InstDrive: Instance-Aware 3D Gaussian Splatting for Driving Scenes}

\author{%
  Hongyuan Liu
  \quad
  Haochen Yu
  \quad
  Bochao Zou
  \quad
  Jianfei Jiang
  \\
  \textbf{Qiankun Liu}
  \quad
  \textbf{Jiansheng Chen}
  \quad
  \textbf{Huimin Ma}
  \vspace{2mm}
  \\
  University of Science and Technology Beijing
}

\begin{document}

\maketitle

\begin{abstract}

Reconstructing dynamic driving scenes from dashcam videos has attracted increasing attention due to its significance in autonomous driving and scene understanding. While recent advances have made impressive progress, most methods still unify all background elements into a single representation, hindering both instance-level understanding and flexible scene editing. Some approaches attempt to lift 2D segmentation into 3D space, but often rely on pre-processed instance IDs or complex pipelines to map continuous features to discrete identities. Moreover, these methods are typically designed for indoor scenes with rich viewpoints, making them less applicable to outdoor driving scenarios. In this paper, we present InstDrive, an instance-aware 3D Gaussian Splatting framework tailored for the interactive reconstruction of dynamic driving scene. We use masks generated by SAM as pseudo ground-truth to guide 2D feature learning via contrastive loss and pseudo-supervised objectives. At the 3D level, we introduce regularization to implicitly encode instance identities and enforce consistency through a voxel-based loss. A lightweight static codebook further bridges continuous features and discrete identities without requiring data pre-processing or complex optimization. Quantitative and qualitative experiments demonstrate the effectiveness of InstDrive, and to the best of our knowledge, it is the first framework to achieve 3D instance segmentation in dynamic, open-world driving scenes.More visualizations are available at our \href{https://instdrive.github.io/}{project page}.

\end{abstract}

\section{Introduction}
Reconstructing dynamic driving scenes from dashcam videos has emerged as a crucial task in autonomous driving, enabling scene-level understanding, behavior prediction, and interactive planning. While recent advances in neural rendering and 3D scene reconstruction—particularly those based on 3D Gaussian Splatting—have shown impressive results in modeling open-world outdoor environments \cite{blocknerf, nsg, unisim, hugs, streetgs}, they largely focus on holistic scene reconstruction without explicit instance-level representations. This limitation hinders downstream applications such as object-level editing, controllable simulation, and fine-grained semantic understanding.

Achieving instance-aware 3D scene reconstruction in driving scenes remains a challenging problem. Compared to indoor scenarios, outdoor driving environments present unique difficulties: long videos with sparse viewpoints, frequent occlusions, multiple moving objects, and significant domain gaps across scenes. Moreover, obtaining ground-truth 3D instance labels is costly and impractical at scale. As a result, recent research has attempted to lift 2D segmentation results into the 3D space. These works can be categorized into three classes:

\begin{figure*}[!htbp]
  \centering
  \includegraphics[width=1.0\textwidth]{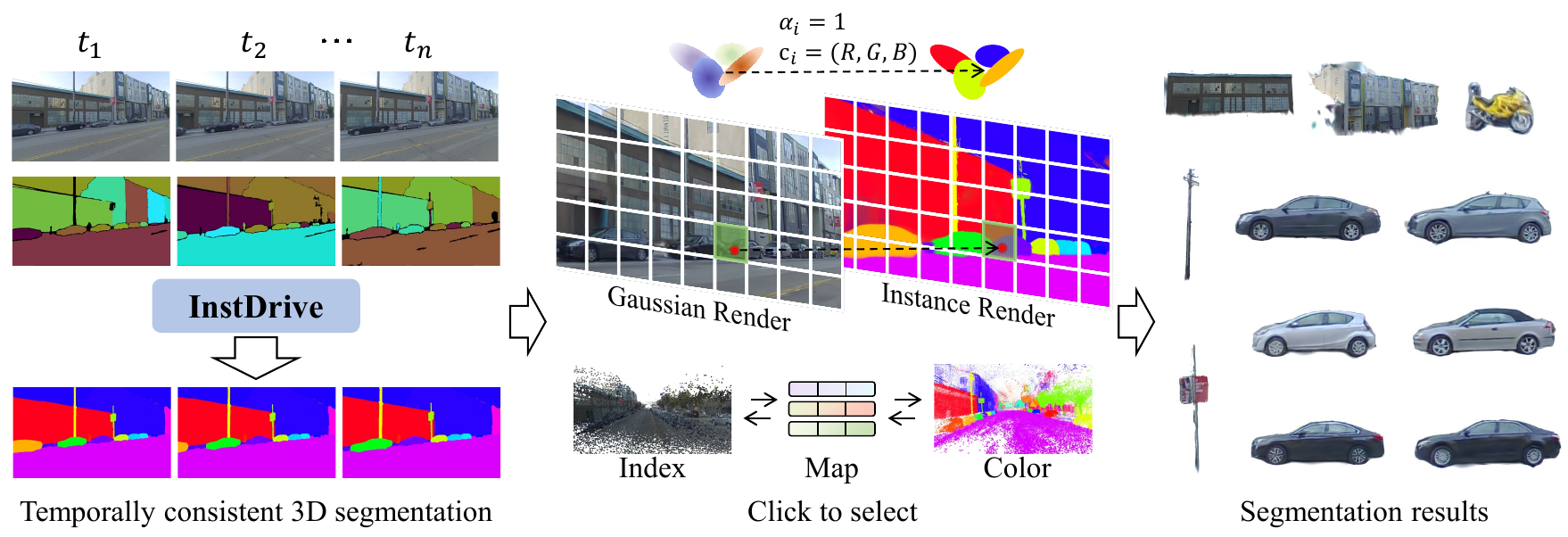}
  \caption{Existing reconstruction methods fail to achieve structured 3D reconstruction with instance-level editability in dynamic driving scenes. To address this, we propose InstDrive, which directly supervises training using SAM-processed video frames without requiring instance matching. We employ a shared 2D–3D color map to enable bijection between instance IDs and colors. During real-time rendering, trained Gaussians are assigned full opacity and colored according to their instance IDs. By capturing click events in the pixel space and retrieving the corresponding color, we map it back to the instance ID using the color map and select all Gaussians associated with that ID, enabling real-time, interactive selection and manipulation of 3D Gaussian instances.} 
  \label{fig:fig1}
\end{figure*}

The first class learns continuous per-point or per-Gaussian features, followed by post clustering (e.g., cosine similarity thresholding) to assign instance labels \cite{lerf, language, langsplat}. However, such methods require sensitive hyperparameter tuning, suffer from limited clustering precision, and often fail to produce globally consistent instance IDs. The second category utilizes pre-processing techniques such as SAM-based tracking and ID propagation \cite{deva} to provide consistent instance IDs across frames \cite{gsgroup}. These methods are highly dependent on tracking quality and often fail under multi-camera view changes or extended video sequences, which are common in autonomous driving scenarios. The third line of work explores codebook-based vector quantization, which maps continuous features to discrete instance identities through learned codebooks \cite{opengs}. These methods eliminate the need for pre-/post-processing and hold promise for instance labeling. Nevertheless, designing a robust and efficient codebook remains a fundamental challenge.

We present InstDrive, a novel instance-aware 3D Gaussian Splatting framework for reconstructing driving scenes with explicit instance-level representations. To ensure high-quality instance feature learning, we introduce a two-stage training strategy that first encourages consistent and discriminative continuous features, followed by efficient and stable instance encoding.

In the continuous feature learning stage, we enforce 2D–3D consistency constraints to guide the instance representation. In 2D space, we utilize SAM \cite{sam} masks to supervise contrastive learning: features within the same instance are pulled closer, while features across different instances are pushed apart. Meanwhile, in 3D space, we partition the point field into voxels and apply a voxel-based spatial consistency loss to mitigate the feature ambiguity introduced by the volumetric nature of Gaussian rendering.

In the instance learning stage, existing codebook-based methods typically rely on scene-specific embeddings, requiring complex clustering and extensive hyperparameter tuning. In category-free instance segmentation, however, where the goal is to assign consistent identities rather than semantic labels, such codebooks often capture redundant variations and lead to uneven latent distributions. To address this, we propose a static, binarized codebook with uniformly distributed codewords fixed throughout training, providing explicit encoding and stabilizing the mapping from features to instance IDs. To mitigate quantization errors, we further introduce a mask-based pseudo-supervision loss, enabling consistent feature encoding and robust supervision without manual labels.

As illustrated in Fig.\ref{fig:fig1}, InstDrive is the first end-to-end framework for scene-level 3D instance reconstruction in open driving environments, requiring only pseudo 2D masks generated by SAM. After reconstruction, it enables real-time, interactive selection of individual instances via simple point-and-click operations within the 3D Gaussian scene. This facilitates a range of downstream editing tasks, including object deletion, color modification, and instance-aware scene manipulation.

\section{Related Work}

\subsection{Neural Rendering for Scene Reconstruction}
Neural rendering has emerged as a key paradigm for 3D scene reconstruction, with Neural Radiance Fields (NeRF) \cite{nerf} pioneering the use of volumetric MLPs for photorealistic novel view synthesis. However, its slow training and rendering speed significantly hinder its applicability in downstream tasks. To address this, many acceleration techniques have been proposed, including hash encoding, coarse-to-fine sampling, and multiscale representations \cite{instantngp, mipnerf, zipnerf}, which have greatly improved rendering efficiency and scalability. Despite these advancements, the implicit volumetric representation of NeRF remains difficult to interpret and edit, limiting its suitability for interactive applications such as autonomous driving simulation. In contrast, 3D Gaussian Splatting \cite{3dgs, 2dgs, gspro} adopts an explicit and compact representation that models the scene as a set of anisotropic Gaussians. This structure enables real-time rendering while maintaining competitive visual quality. Given the strong real-time requirements of driving scenes and the need for editable 3D representations, we adopt 3D Gaussian Splatting as the foundational representation in our framework.

\subsection{Reconstruction of Driving Scenes}

Neural rendering has become central to autonomous driving scene reconstruction, enabling photorealistic digital twins from real-world sensor data. Extensions of NeRF to driving scenarios address challenges such as dynamic objects, large-scale environments, and multi-sensor inputs. For instance, NSG \cite{nsg} models multi-object dynamics via scene graphs for object-level manipulation; MARS \cite{mars} builds a modular, instance-aware NeRF that separates dynamic actors from static backgrounds; and UniSim \cite{unisim} constructs neural feature grids from driving logs to support multi-sensor simulation and controllable scene synthesis. To further improve real-time performance and editability, 3D Gaussian-based methods have gained traction. Street Gaussians \cite{streetgs} use semantic Gaussian primitives with tracked poses for real-time rendering and composable street scenes. HUGS \cite{hugs} and OmniRe \cite{omnire} enhance scalability and interactivity in urban reconstructions through advanced 3DGS techniques. However, most existing approaches still rely on unified scene representations or semantic segmentation, lacking explicit 3D instance-level decomposition. This limits fine-grained understanding and flexible manipulation of individual elements—capabilities that are essential for downstream tasks like autonomous driving, simulation, and digital twin construction.

\subsection{3D Instance Segmentation via 2D-to-3D Lifting}

Understanding 3D scenes at the instance level is fundamental for many applications, yet remains challenging due to the high cost of acquiring dense and diverse 3D annotations. To address this, recent works have increasingly explored how to lift pre-trained 2D vision and vision-language models to enable efficient 3D scene understanding. This 2D-to-3D lifting paradigm allows models to leverage abundant 2D data and supervision while minimizing reliance on expensive 3D labels.

Recent studies on point cloud understanding have explored lifting 2D visual and linguistic priors into 3D via open-vocabulary learning. By aligning 2D foundation models (e.g., CLIP \cite{clip}, DINO \cite{dino}) with point cloud features through contrastive training or multi-view fusion, these methods enable zero-shot 3D recognition and segmentation without 3D annotations \cite{ulip, clip2point, pointclipv2, lu2023open}.

NeRF have been extended beyond photorealistic rendering to enable semantic scene understanding by incorporating semantics into implicit 3D representations. Several approaches adapt 2D segmentation models such as SAM \cite{sam} into 3D by leveraging density-guided projection and cross-view fusion, enabling promptable segmentation and hierarchical grouping in 3D space \cite{sa3d, san, garfield}. The explicit representation of 3DGS enables semantic extensions \cite{langsplat, language, featuregs}, but most methods struggle with consistent instance ID mapping across views. Gaussian Grouping \cite{gsgroup} and CoSSegGaussians \cite{cosseggaussians} achieve global consistency through pre-processing, yet fail to generalize to long, multi-view driving sequences. OmniSeg3D \cite{omniseg3d} and OpenGaussian \cite{opengs} use contrastive learning without tracking data, but their complex pipelines limit scalability in open driving scenarios.

\begin{figure*}[!t]
  \centering
  \includegraphics[width=1.0\textwidth]{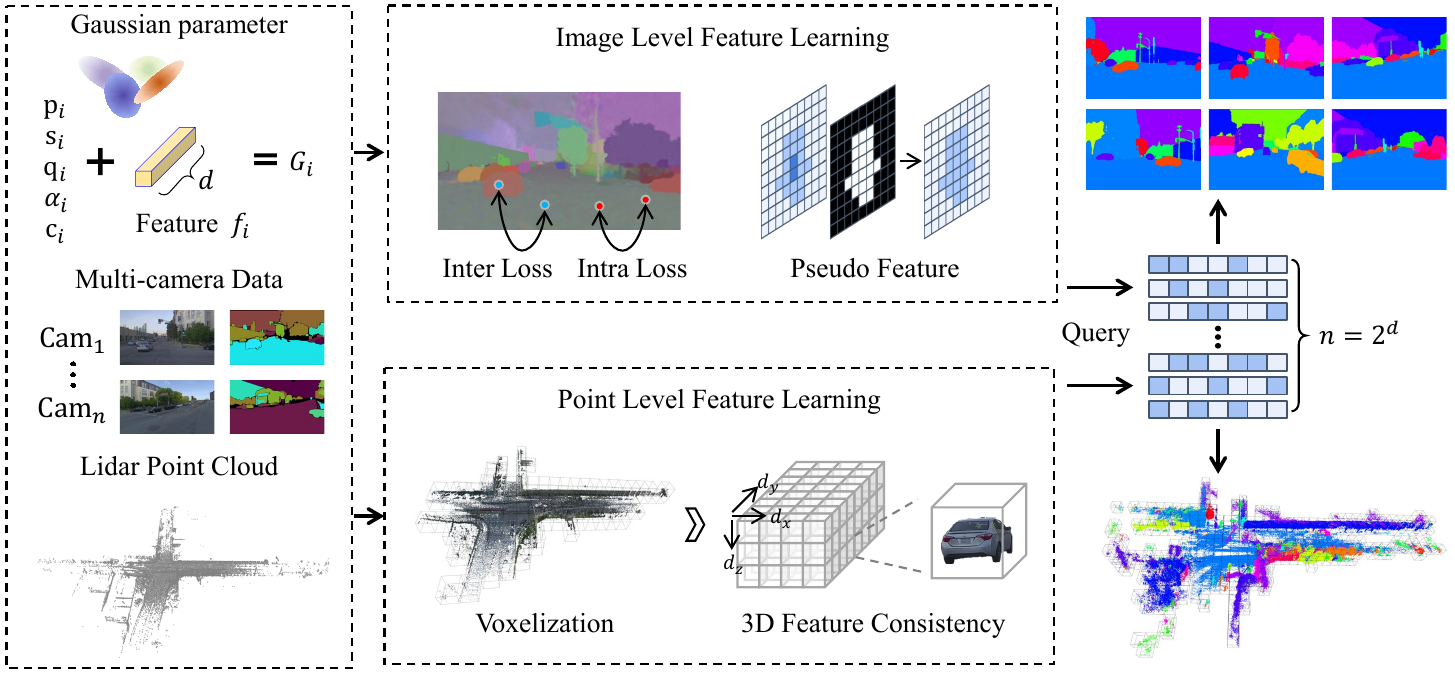}
  \caption{Framework Overview. We extend Gaussian attributes with an instance feature dimension and train the scene using multi-view images and LiDAR points. A contrastive loss and a voting-based pseudo-supervision loss guide 2D feature learning, while a voxel-based consistency loss enforces 3D coherence by aligning nearby Gaussians. Both 2D and 3D features are mapped to discrete instance IDs via a binarized static codebook.} 
  \label{fig:frame}
\end{figure*}

\section{Methodology}

\subsection{Preliminaries}
3D Gaussian Splatting (3DGS) \cite{3dgs}provides a compact, explicit, and differentiable representation of 3D scenes, making it particularly suitable for instance-level understanding tasks. By associating each Gaussian with localized geometry, appearance, and additional learnable attributes, 3DGS naturally supports structured reasoning at the object level. This per-primitive expressiveness enables fine-grained semantic modeling and facilitates integration with point-based or region-based supervision signals. Gaussian Splatting represents a 3D scene as a set of $N$ anisotropic 3D Gaussians $\mathcal{G} = \{g_i\}_{i=1}^{N}$, where each Gaussian is parameterized as $g_i = (p_i, s_i, q_i, \alpha_i, c_i )$, with 3D position $p_i \in \mathbb{R}^3$, scale $s_i \in \mathbb{R}^3$, orientation represented by a unit quaternion $q_i \in \mathbb{R}^4$, opacity $\alpha_i \in [0, 1]$, and color $c_i$ modeled using spherical harmonics (SH) coefficients. For each pixel $u$, the set of overlapping Gaussians $\mathcal{N}_{u}$ is computed and sorted in front-to-back order. The final color $\mathbf{c}_{u}$ at pixel $u$ is composited using alpha blending:
\begin{equation}
\label{eq:color_blend}
\mathbf{c}_{u} = \sum_{i \in \mathcal{N}_{u}} c_i \cdot \alpha'_i \cdot \prod_{j=1}^{i-1} (1 - \alpha'_j),
\end{equation}
where $\alpha'_i = \alpha_i \cdot G_i'(u)$, and $G_i'(u)$ denotes the 2D projected Gaussian weight at pixel $u$. To enable semantic and instance-level scene understanding, we augment each Gaussian with a learnable feature vector $f_i \in \mathbb{R}^d$. The rendering of the feature $\mathbf{f}_{u}$ follows the same compositing rule as color. This differentiable feature-aware Gaussian representation forms the foundation of our framework, enabling structured perception in complex, dynamic driving environments.

\subsection{Continuous Instance Feature Learning}
\paragraph{Contrastive Learning for Instance-Level Features}

When rendering the scene from a given viewpoint, each 3D Gaussian feature is projected into a 2D feature map, alongside the RGB image, such that each pixel $u$ obtains a feature vector $\mathbf{f}_u$. These 2D features are used to supervise the 3D representation via differentiable rendering. Since we do not perform cross-frame instance tracking and rely solely on per-frame masks without category labels, we lack ground-truth instance IDs across views. We adopt the contrastive loss formulation introduced in prior work \cite{omniseg3d, contrastive, opengs}, leveraging its established effectiveness in promoting instance-aware feature clustering. The core insight is that pixels rendered from 3D Gaussians and falling within the same mask region should share the same object identity and thus have similar features, whereas features from different masks should be distinct to improve instance-level separability. Let $M^1, M^2, \dots, M^N$ denote the $N$ segmentation masks generated by SAM for a given training image. For each mask $M^i$, we compute its prototype feature $\bar{\mathbf{f}}^i$ by taking the average of the features of all pixels within that region. We then define a unified contrastive loss that combines both intra-mask consistency and inter-mask discrimination:

\begin{equation}
\mathcal{L}_{contra} = 
\underbrace{
\frac{1}{N} \sum_{i=1}^{N} \frac{1}{|M^i|} \sum_{u \in M^i} \big\|\mathbf{f}_u - \bar{\mathbf{f}}^i \big\|^2
}_{\text{Intra-mask}} 
+
\underbrace{
\frac{1}{N(N-1)} \sum\limits_{i=1}^{N} \sum\limits_{j=1, j\neq i}^{N} \frac{1}{ \big\|\bar{\mathbf{f}}^i - \bar{\mathbf{f}}^j\big\|^2}
}_{\text{Inter-mask}}
\label{inter_intra}
\end{equation}

The Intra-mask part encourages all features within a mask to be close to their prototype, enforcing intra-instance smoothness and coherence. The Inter-mask part penalizes similarity between prototype features of different masks, promoting clear separation in feature space. This unified loss supervises the 3DGS model to generate instance-aware embeddings without requiring any multi-view instance matching. It ensures that pixels belonging to the same mask are pulled together in the feature space, while those from different masks are pushed apart, improving the discriminability and structure of learned 2D features.

\paragraph{Voxel-Based Feature Consistency}

Although differentiable rendering in 3DGS enables alignment with 2D instance masks, training instance features solely with 2D supervision often leads to overfitting, failing to capture coherent 3D representations. This is due to the volumetric nature of rendering, where each pixel aggregates features from multiple Gaussians along the viewing ray—allowing groups of Gaussians to explain 2D masks without consistent 3D embeddings. To address this, we introduce a voxel-based consistency loss that encourages nearby Gaussians to share similar instance features, based on the intuition that spatial proximity implies semantic similarity. We partition the scene into a regular voxel grid and align each Gaussian’s feature with the voxel's mean feature. To avoid semantic mixing at object boundaries caused by fixed voxelization, we apply a random voxel shift strategy: the grid is perturbed by offsets randomly sampled within one voxel length along each axis. This introduces stochasticity in voxel assignments and improves the robustness of 3D spatial supervision. Formally, the voxel-based consistency loss $\mathcal{L}_{\text{voxel}}$ is defined as:

\begin{equation}
\mathcal{L}_{\text{voxel}} = \frac{1}{|V|} \sum_{v \in V} \frac{1}{|\mathcal{G}_v|} \sum_{g \in \mathcal{G}_v} \big\| \mathbf{f}_g - \bar{\mathbf{f}}_v \big\|^2,
\end{equation}

where $\mathbf{f}_g$ is the instance feature of Gaussian $g$, and $\bar{\mathbf{f}}_v$ is the mean feature of all Gaussians within voxel $v$. This loss encourages Gaussians that are spatially close (under a jittered voxel partition) to share consistent instance-level features. When combined with 2D contrastive loss, it enables the model to learn embeddings that are both view-aware and spatially coherent in 3D space.

\begin{figure*}[!t]
  \centering
  \includegraphics[width=1.0\textwidth]{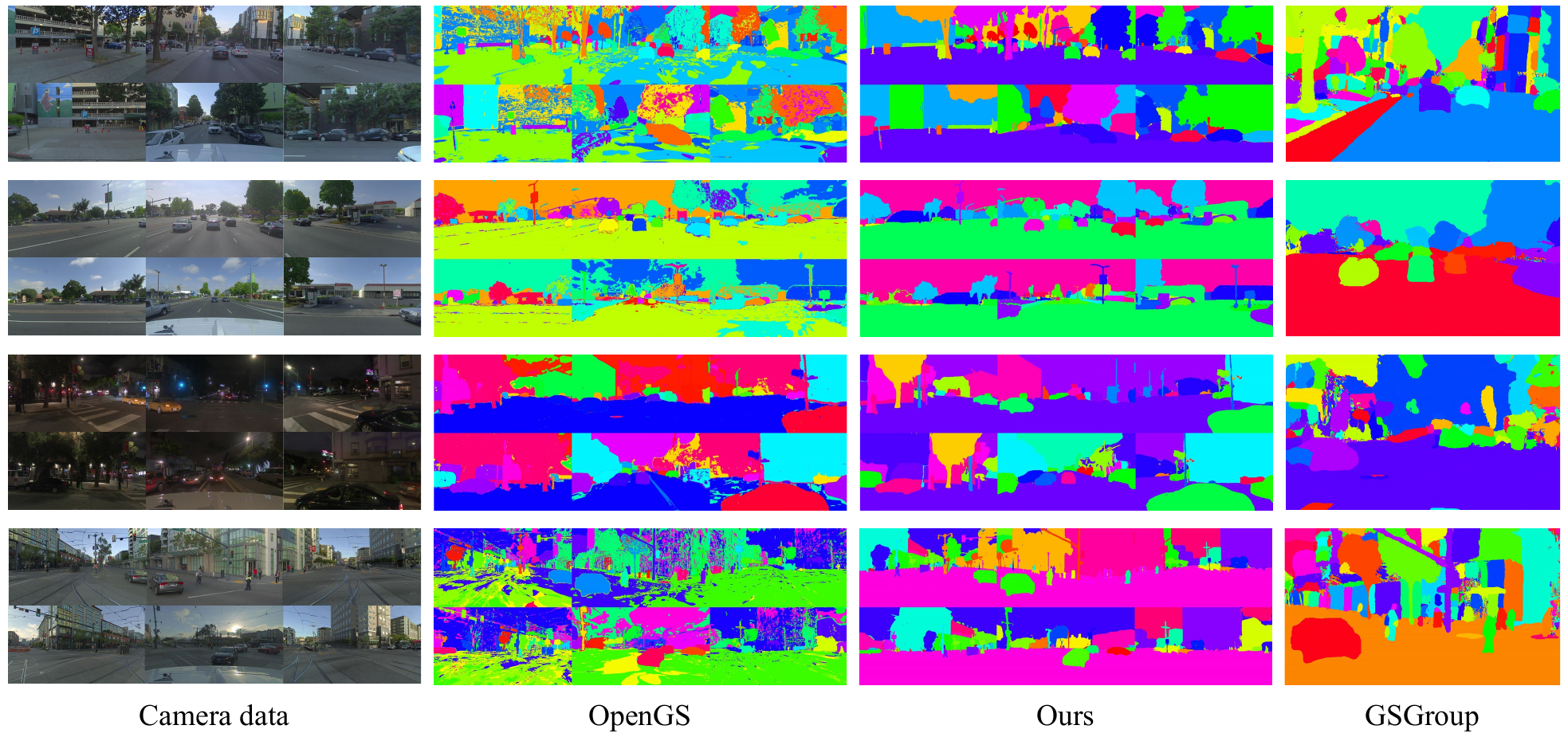}
  \caption{2D segmentation results. We present 2D segmentation results across diverse scenes and conditions. The first column shows six-view camera inputs arranged in a 2×3 grid (from top to bottom, left to right): front-left, front, front-right, left, back, and right views. The second, third, and fourth columns display results from OpenGS, our method, and GSGroup, respectively. Due to preprocessing failures when GSGroup switches between multiple camera views, we only include its result under the forward-facing view. As shown in the visualizations, our method produces consistent multi-view segmentations, whereas baseline methods often exhibit distortions and significant noise (best viewed in color and zoomed in).} 
  \label{fig:feat2d}
\end{figure*}

\subsection{Quantized Instance Feature Learning}
\paragraph{Instance Quantization via Static Codebook}

The learned feature representations are typically continuous, which are effective for general tasks but ill-suited for discrete instance-level decisions. Without ground-truth instance IDs, the challenge lies in partitioning these features into distinct, instance-aware clusters. A common approach is to compare query features with all instance features using metrics like cosine similarity, followed by threshold-based filtering. However, this method cannot produce global instance segmentation and is highly sensitive to threshold selection. Vector quantization offers an alternative solution. Early applications in 3DGS focus on compression, grouping similar Gaussians into shared codebook vectors to reduce memory and computation \cite{compgs, lightgs}. Recent work has explored its use for semantic understanding \cite{opengs}, but challenges remain: codebook updates require costly clustering (e.g., k-means), and learned vectors often lack uniformity, leading to overlapping clusters, unstable training, and ambiguous instance assignments.

Our key insight is that, unless semantic attributes are explicitly required, the sole purpose of instance features is to provide a unique identifier for each instance. Based on this, we propose a static codebook design that eliminates redundant learning and ensures uniformly distributed cluster centers in feature space. Let each instance feature be a $d$-dimensional vector. The codebook is constructed by taking the Cartesian product, resulting in a static codebook $\mathcal{C} = \{\mathbf{c}_1, \mathbf{c}_2, \dots, \mathbf{c}_{2^d}\}$ of $2^d$ binary vectors, where each $\mathbf{c}_i \in \{-1,1\}^d$. The instance quantization process begins by applying a $\tanh$ activation to the learned 3D or 2D instance feature $\mathbf{f} \in \mathbb{R}^d$, yielding a bounded vector $\hat{\mathbf{f}} = \tanh(\mathbf{f})$ within the range $(-1, 1)$. We then identify its nearest codebook entry by minimizing the L2 distance:

\begin{equation}
    \text{ID}(\mathbf{f}) = \arg\min_{i} \big\| \hat{\mathbf{f}} - \mathbf{c}_i \big\|_2^2 
    \label{eq:query}
\end{equation}

This quantization scheme enforces an explicit and consistent mapping from continuous instance features to discrete codebook IDs. The use of a static, binary codebook brings several advantages: no additional parameters or updates are required during training, the cluster centers are evenly distributed, and the distances between them remain fixed throughout learning. By integrating this quantization mechanism, the model is able to produce globally unique instance identifiers directly from learned features, without the need for post-processing or threshold tuning. This enables efficient and end-to-end 3D instance segmentation purely through differentiable learning.


\begin{figure*}[!t]
  \centering
  \includegraphics[width=1.0\textwidth]{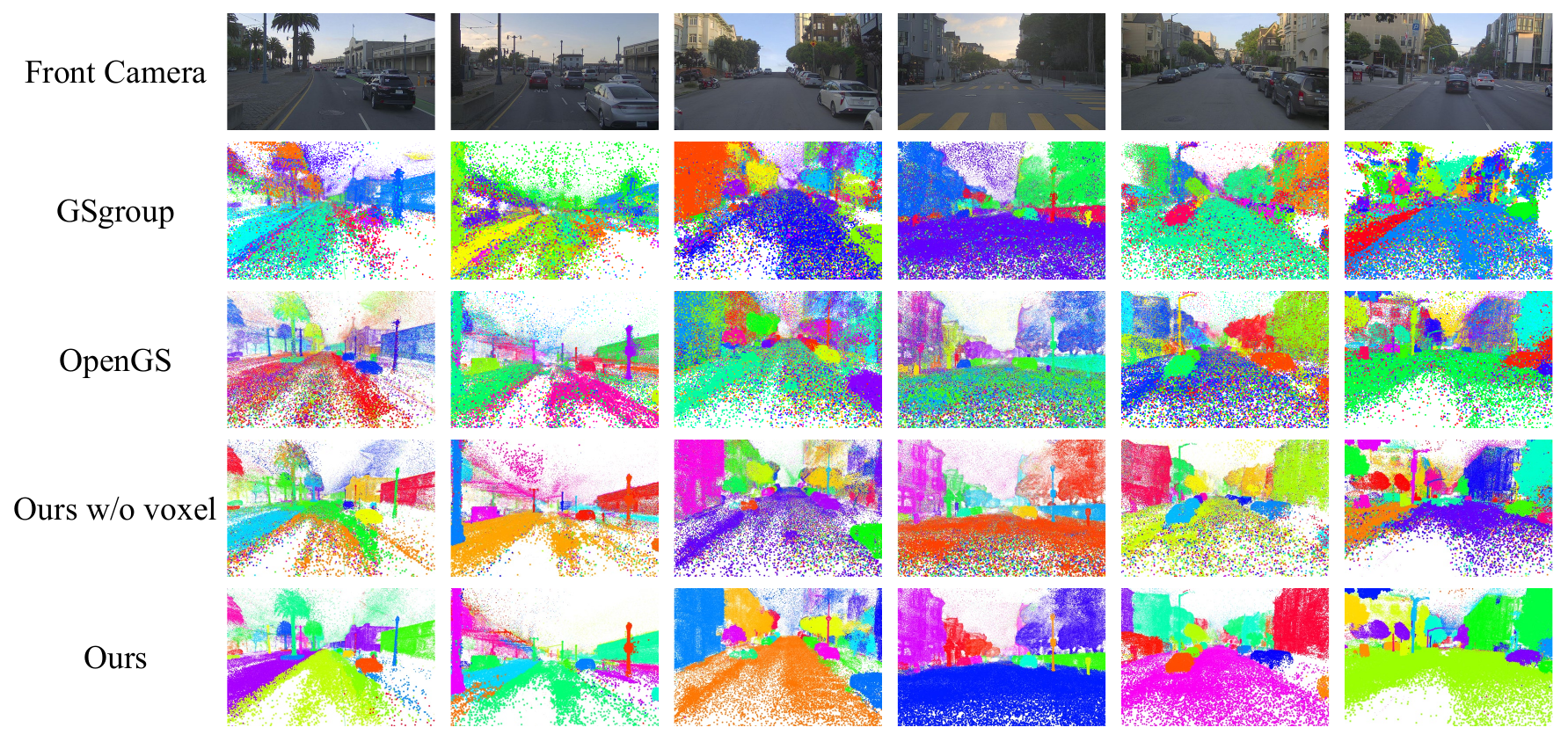}
  \caption{Comparison of point-level instance segmentation results. The first row shows the forward-facing camera image of the corresponding scene, followed by results from GSGroup, OpenGS, our method without voxel loss, and the full version of our method. As observed from the visualizations, baseline methods often produce fragmented segmentation with noisy points within the same instance. In contrast, our method yields more coherent results, and the incorporation of voxel loss further enhances 3D consistency, resulting in cleaner and more complete instance masks.} 
  \label{fig:point}
\end{figure*}

\paragraph{Instance-Consistent Pseudo Supervision}

While the contrastive learning objective defined in Eq.\ref{inter_intra} is effective in the early stages of training for learning 2D instance-aware features, it often fails to provide sufficient gradients as learning progresses. To further enhance the spatial coherence of the projected 2D features and produce more complete and consistent instance segmentation, we introduce an instance-consistent pseudo supervision mechanism based on majority voting. 

Given a rendered 2D feature map $\mathbf{F} \in \mathbb{R}^{H \times W \times d}$, we first apply vector quantization to each pixel-wise feature using the static codebook, following Eq.\ref{eq:query}. This yields a quantized 2D instance ID map, where each pixel is assigned the index of its nearest codebook vector. We then apply majority voting within each SAM-provided mask region: for a given mask $M^i$, we count the occurrence of each quantized ID within the mask and assign the most frequent ID as the representative instance label for the entire mask. All features within $M^i$ are then replaced with the corresponding codebook vector, resulting in a pseudo supervision feature map $\mathbf{f}^\text{pseudo}$. The pseudo supervision loss $\mathcal{L}_\text{pseudo}$ is defined as:
\begin{equation}
\mathcal{L}_{\text{pseudo}} = \frac{1}{|\Omega|} \sum_{u \in \Omega} \big\| \mathbf{f}_u - \mathbf{f}_u^\text{pseudo} \big\|^2,
\end{equation}

where $\Omega$ denotes the set of all valid pixels in the image, $\mathbf{f}_u$ is the original rendered feature at pixel $u$, and $\mathbf{f}_u^\text{pseudo}$ is the pseudo-supervised feature obtained through the voting process. This voting-based supervision helps to eliminate fragmented or inconsistent predictions within each mask, reinforcing feature consistency at the instance level. Unlike directly using in-mask feature averages, which may introduce ambiguity at object boundaries, the use of quantized codebook vectors provides a clear and discrete supervision signal.

\begin{figure*}[!t]
  \centering
  \includegraphics[width=1.0\textwidth]{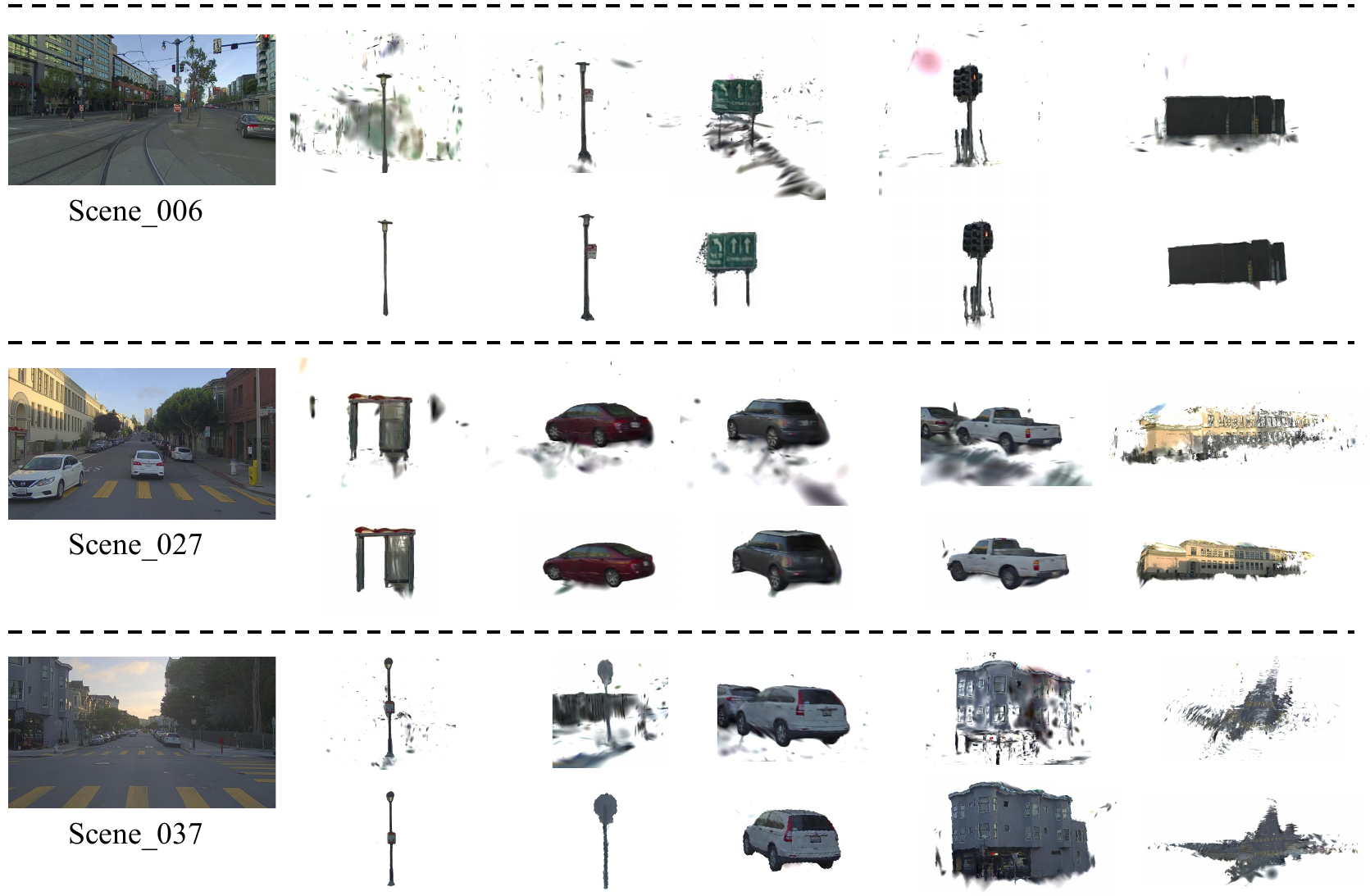}
  \caption{3D instance segmentation results. Due to GSGroup’s limited applicability, we compare only OpenGS and our method. The first and second rows show results from OpenGS and our method, respectively. Each region represents an instance, with overlaps indicating instance coupling. We present results across diverse scenes and object scales, including street lamps, traffic signs, buildings, and road surfaces. Our method produces more complete segmentations with less scattered points and artifacts, while OpenGS suffers from noise and instance entanglement.} 
  \label{fig:click}
\end{figure*}

\section{Experiment}

\subsection{Settings}
\paragraph{Datasets and Metrics}
We conduct all experiments on full PandaSet \cite{pandaset}, a large-scale multimodal dataset for autonomous driving that provides synchronized LiDAR and camera data with high-quality 3D annotations. It contains 103 scenes with diverse traffic conditions. For evaluation, we follow prior works \cite{opengs, gsgroup} and adopt mean Intersection over Union (mIoU) and mean Accuracy (mAcc) as the primary metrics across all classes. However, as both metrics are computed on projected 2D masks, higher scores do not necessarily indicate better 3D instance segmentation performance—thus, we also refer readers to our 3D qualitative results for a more comprehensive assessment.

\paragraph{Implementation Details}
All experiments are conducted on a single NVIDIA RTX 4090 GPU. For a fair comparison, we follow the original GSGroup \cite{gsgroup} setup by using DEVA \cite{deva} to preprocess instance masks. In contrast, both our method and OpenGS \cite{opengs} directly utilize masks generated by SAM without additional preprocessing. In our implementation, the feature dimension for each 3D Gaussian instance is set to 8. We train all scenes for 30,000 iterations, with the codebook module activated starting from the 10,000th iteration. To improve computational efficiency, the voxel-based consistency loss is computed every five iterations using a voxel grid size of 0.5 meters. We set the weights of the $\mathcal{L}_{\text{voxel}}$, $\mathcal{L}_{contra}$, $\mathcal{L}_{\text{pseudo}}$ to 0.1.

\subsection{Quantitative Comparison}

The quantitative results are presented in Tab.\ref{tab:pandast_results}. On the PandaSet dataset, we evaluate performance under three different camera configurations: 1Cam (forward-facing only), 3Cam (forward, front-left, and front-right), and 6Cam (all available cameras). Notably, GSGroup relies on tracking-based preprocessing, which fails under multi-view setting due to inconsistent instance association across views. Therefore, we only report its result under the 1Cam configuration, highlighting the limitations of preprocessing-dependent approaches.

Our method consistently outperforms OpenGS across all configurations and achieves competitive performance with GSGroup under the 1Cam setting, despite using no tracking and relying solely on SAM-generated masks. It is worth noting that the reported mIoU and mAcc are 2D metrics and do not fully reflect the quality of 3D instance segmentation; we refer readers to our qualitative results for a more comprehensive comparison.

\subsection{Qualitative Comparison}
We present a qualitative comparison across 2D segmentation, point clouds, and 3D Gaussian instances. As shown in Fig.\ref{fig:feat2d}, instance masks are rendered from 2D features, with each instance assigned a unique color. Results for OpenGS and our method are shown across six camera views, while GSGroup is limited to the forward-facing view due to failure under multi-view setting. Compared to OpenGS—which suffers from heavy noise in segmentation, especially on large structures like the ground and buildings—our method achieves cleaner, more view-consistent results. Additionally, it outperforms GSGroup in preserving object shapes and handling multi-view data without ground-truth instance supervision.

We map the instance features of all points to unique IDs and colors, yielding 3D point cloud segmentation results, as shown in Fig.\ref{fig:point}. While GSGroup provides seemingly valid 2D segmentations, its 3D results are noisy and fail to properly segment 3D Gaussian objects. OpenGS faces similar issues, particularly with large-scale objects like ground and buildings. In contrast, our method provides consistent and clear 3D segmentation. As demonstrated in Fig.\ref{fig:click}, OpenGS is prone to merging different instance Gaussian points due to noise in small-scale objects. For large-scale objects, the lack of global consistency leads to the splitting of the same object into multiple instances. Our approach consistently achieves stable, accurate results across all scenarios.

\begin{table}[t]
\centering
\caption{Comparison of segmentation performance across different camera setups on the full Pandast dataset.}
\label{tab:pandast_results}
\begin{tabular}{lcccccc}
\toprule
\multirow{2}{*}{Method} & \multicolumn{2}{c}{1cam} & \multicolumn{2}{c}{3cam} & \multicolumn{2}{c}{6cam} \\
\cmidrule(lr){2-3} \cmidrule(lr){4-5} \cmidrule(lr){6-7}
& mIoU & mAcc & mIoU & mAcc & mIoU & mAcc \\
\midrule
gsGroup & \textbf{0.7694} & 0.8606 & -      & -      & -      & -      \\
openGS  & 0.6505 & 0.8124 & 0.6318 & 0.7833 & 0.5665 & 0.7274 \\
ours    & 0.7535 & \textbf{0.9022} & \textbf{0.7410} & \textbf{0.8853} & \textbf{0.7091} & \textbf{0.8577} \\
\bottomrule
\end{tabular}
\end{table}

\begin{table}[!t]
\centering
\begin{minipage}[t]{0.48\linewidth}
\centering
\caption{Voxel and Pseudo loss ablation.}
\label{tab:ablation_loss}
\begin{tabular}{ccccc}
\toprule
\textbf{Case} & \texttt{Voxel} & \texttt{Pseudo} & \textbf{mIoU} & \textbf{mAcc} \\
\midrule
Full       & \checkmark & \checkmark & 0.7091 & 0.8577 \\
w/o Voxel & ×          & \checkmark & \textbf{0.7117} & \textbf{0.8602} \\
w/o Pseudo & \checkmark & ×          & 0.6724 & 0.8173 \\
\bottomrule
\end{tabular}
\end{minipage}
\hfill
\begin{minipage}[t]{0.48\linewidth}
\centering
\caption{Instance feature dimension ablation.}
\label{tab:ablation_dim}
\begin{tabular}{ccc}
\toprule
\textbf{Dim} & \textbf{mIoU} & \textbf{mAcc} \\
\midrule
6  & 0.6826 & 0.8448 \\
8  & 0.7091 & 0.8577 \\
10 & \textbf{0.7192} & \textbf{0.8634} \\
\bottomrule
\end{tabular}
\end{minipage}
\end{table}

\subsection{Ablation Study}

As shown in Tab.\ref{tab:ablation_loss}, we conduct ablation studies on the two proposed loss functions. The pseudo-supervised loss significantly improves the quality of 2D segmentation. For the voxel-based 3D consistency loss, we observe a slight improvement in 2D performance when it is removed. However, our primary objective is to achieve spatially consistent 3D segmentation. As illustrated in Fig.\ref{fig:point}, the proposed consistency loss effectively suppresses intra-instance noise and enhances the overall quality of 3D segmentation. Therefore, we retain this loss term in our final model.

We also perform ablation studies on the dimensionality of Gaussian features, as shown in Tab.\ref{tab:ablation_dim}. Increasing the feature dimension leads to moderate improvements in 2D segmentation performance. However, a higher dimensionality results in an exponential increase in both codebook size and computational cost. In practice, we find that a feature dimension of 8 with a codebook size of 256 is sufficient. For fair comparison and computational efficiency, we adopt a feature dimension of 8 in all experiments.

\section{Conclution}
We propose InstDrive, the first end-to-end framework for instance-aware 3D Gaussian Splatting in dynamic driving scenes. Unlike prior methods relying on tracking or clustering, InstDrive directly reconstructs editable instance-level representations from dashcam videos and pseudo 2D masks. Our two-stage training strategy combines 2D–3D consistency constraints with a static binarized codebook to ensure robust and efficient instance encoding. Without requiring manual labels, our method achieves consistent 3D segmentation and supports real-time, point-and-click instance editing. Experiments on the full PandaSet dataset validate the effectiveness of InstDrive in producing high-quality, multi-view instance reconstructions, offering a practical solution for scene-level understanding and interactive applications in autonomous driving.

\paragraph{Limitations}


(1) The codebook capacity is manually set and may not adapt well to varying object counts across scenes. A fixed-size codebook can lead to either insufficient IDs or over-segmentation. Future work will explore dynamic codebook mechanisms for better scene adaptability. (2) At the current stage, the learned instance features do not contain semantic information. A potential future direction is to distill semantics from pre-trained models into the Gaussian instance features, enabling more comprehensive scene understanding.

\bibliographystyle{IEEEtran}
\bibliography{neurips_2025}

\end{document}